\pdfoutput=1

\documentclass[11pt]{article}

\usepackage[]{acl}

\usepackage{times}
\usepackage{latexsym}

\usepackage[T1]{fontenc}

\usepackage[utf8]{inputenc}

\usepackage{microtype}

\usepackage{inconsolata}

\usepackage{microtype}
\usepackage{inconsolata}
\usepackage{amsmath,amsfonts}
\usepackage{multirow,subfigure,caption,amssymb,booktabs,xcolor}
\usepackage{graphicx}
\usepackage{array}
\newcommand{\modelname}{UniPCM} 

\newcommand{\dataname}{UniPreDial}

\newcommand{\methodname}{TAP} 

\newcommand{\pet}{PET} 

\usepackage{fancyhdr}
\title{UniPCM: Universal Pre-trained Conversation Model with Task-aware 
Automatic Prompt}

 \author{Yucheng Cai \thanks{$^{*}$the work was done during internship at DAMO Academy} $^{,1}$\\
   \\\And
   , Wentao Ma $^{2}$\\
  \\\And
  , Yuchuan Wu $^{2}$\\
  \\\And
  , Shuzheng Si $^{2}$ \\
    $^{1}$Speech Processing and Machine Intelligence (SPMI) Lab, Tsinghua University, Beijing, China \\
 $^{2}$ DAMO Academy, Alibaba Group \\
 \texttt{cyc22@mails.tsinghua.edu.cn}
  \\\And
  , Yuan Shao $^{2}$\\
  \\\And
  , Zhijian Ou $^{1}$\\
  \\\And
  , Yongbin Li $^{2}$\\
  }

\begin{document}

\maketitle

\begin{abstract}

Recent research has shown that multi-task pre-training greatly improves the model's robustness and transfer ability, which is crucial for building a high-quality dialog system. 
However, most previous works on multi-task pre-training rely heavily on human-defined input format or prompt, which is not optimal in quality and quantity. 
In this work, we propose to use \textbf{T}ask-based \textbf{A}utomatic \textbf{P}rompt generation (\methodname{}) to automatically generate high-quality prompts. Using the high-quality prompts generated, we scale the corpus of the pre-trained conversation model to 122 datasets from 15 dialog-related tasks, resulting in \textbf{Uni}versal \textbf{P}re-trained \textbf{C}onversation \textbf{M}odel (\textbf{UniPCM}), a powerful foundation model for various conversational tasks and different dialog systems. 
Extensive experiments have shown that \modelname{} is robust to input prompts and capable of various dialog-related tasks. Moreover, \modelname{} has strong transfer ability and excels at low resource scenarios, achieving SOTA results on 9 different datasets ranging from task-oriented dialog to open-domain conversation. 
Furthermore, we are amazed to find that TAP can generate prompts on par with those collected with crowdsourcing. 
The code is released with the paper. 

\end{abstract}



\section{Introduction}

Recently, dialogue systems have been developing rapidly in various scenarios, such as personal assistants and customer service. 
The advancements in dialogue systems for those applications have been significantly boosted by the use of pre-trained language models (PLMs), including GPT-2 \citep{radford2019gpt2}, BERT \citep{kenton2019bert}, and T5 \citep{raffel2020t5} , combined with task-specific fine-tuning on annotated data \citep{hosseini2020simple, yang2021ubar, heck2020trippy, lee-2021-improving-end, liu2022revisiting}.
 However, most of the models trained under the `pretrain-finetune' paradigm are limited to specific tasks or datasets, and the dialog systems 
 built upon those models can only respond to certain input formats, which lacks robustness and transfer ability.

To relieve such problems, 
 multi-task pre-training, which has achieved great success in language model pre-training, has been introduced to pre-trained conversation models (PCM). 
 Recent progress in multi-task pre-training \citep{ouyang2022training,sanh2022multitask, mishra2022cross, wang2022super} has shown that the robustness and transfer ability of language models are greatly improved by pre-training with multiple tasks. 
 

 However, the previous works on multi-task learning rely heavily on human-defined input format or prompt.
 We find those artificially constructed prompts still have two obvious weaknesses, which can be relieved by our proposed task-aware automatic prompt generation method \methodname{}:
 
(1) \textbf{Human labor required and limited in quantity.} Previous works like Supernatural-instruction \citep{wang2022super} only have one task instruction for each dataset, which is difficult for the model to catch the essence of the tasks and transfer to other prompts. In contrast, our TAP method can generate numerous prompts given a task, and we show in our experiments that increasing the number of prompts not only scales the pre-training corpus, but makes the model understand the task better as well.

(2) \textbf{Hard to understand and limited in quality.} Human labelers easily incorporate their own understandings into the prompts or simply obtain the prompt 
by paraphrasing the dataset description, which makes the prompts quite long and unnatural, containing specific knowledge of the datasets. Meanwhile, our TAP method leverages task-related information to generate task-centric prompts and the quality is ensured by scoring and filtering procedure. The superiority of the generated prompts in quality is proved by both automatic and human evaluation.
The TAP method can generate numerous high-quality prompts, which can greatly help train a universal pre-trained conversational model by scaling the pre-training datasets and fuse different tasks using the proposed multi-prompt training mechanism. Using the 303 high-quality prompts automatically generated for the 15 tasks, we scale our pre-training corpus to 122 datasets and 26,625,486 instances, which, to our knowledge, is currently the largest annotated conversational dataset, covering almost all dialog-related tasks. 
 Moreover, we propose a multi-prompt training mechanism to make use of the generated prompts to better fuse different tasks. The resulting model \modelname{} is a powerful foundation model for various conversational tasks and different dialog systems.

Through comprehensive experiments in multiple conversational tasks, we find that \modelname{} has strong ability on various dialog tasks, which outperforms the T5 baseline by 7.14\%  in the few-shot setting on DialoGLUE \citep{Mehri2020DialoGLUE} and achieves SOTA results on 9 different datasets ranging from task-oriented dialog to open-domain conversation. 
 Moreover, the model is robust to input format and can respond to different input prompts. 
 Furthermore, to comprehensively evaluate the quality of the generated prompts, we conduct human evaluation and automatic evaluation. Our generated prompts achieve higher average scores than human-written prompts by 9.50\%  on three proposed metrics in human evaluation and improve by 2.40\% when used to finetune downstream tasks. 

In summary, our main contributions are:
\begin{itemize}
\item  We propose a task-aware automatic prompt generation method \methodname{} to better fuse the datasets from different tasks in the multi-task pre-training, 
which can generate numerous high-quality prompts based on extracted task information. The proposed method 
 greatly reduces human effort in prompt engineering and improves the quality of generated prompts.
\item Leveraging the high-quality prompts generated, we pre-train a unified pre-trained conversation model UniPCM 
by scaling the  the pre-training datasets to 122 dialog-related datasets from 15 dialog-related tasks, resulting in a powerful conversation model \modelname{}. 
The pre-trained model and the datasets collected will be released to public.

\item We conduct extensive experiments on 10 dialog-related benchmarks including 6 types of task. 
Results on few-shot and full-data experiments show the superiority of our proposed method and model.
\end{itemize}

\begin{figure*}[t]
\centering
	\includegraphics[width=0.98\linewidth]{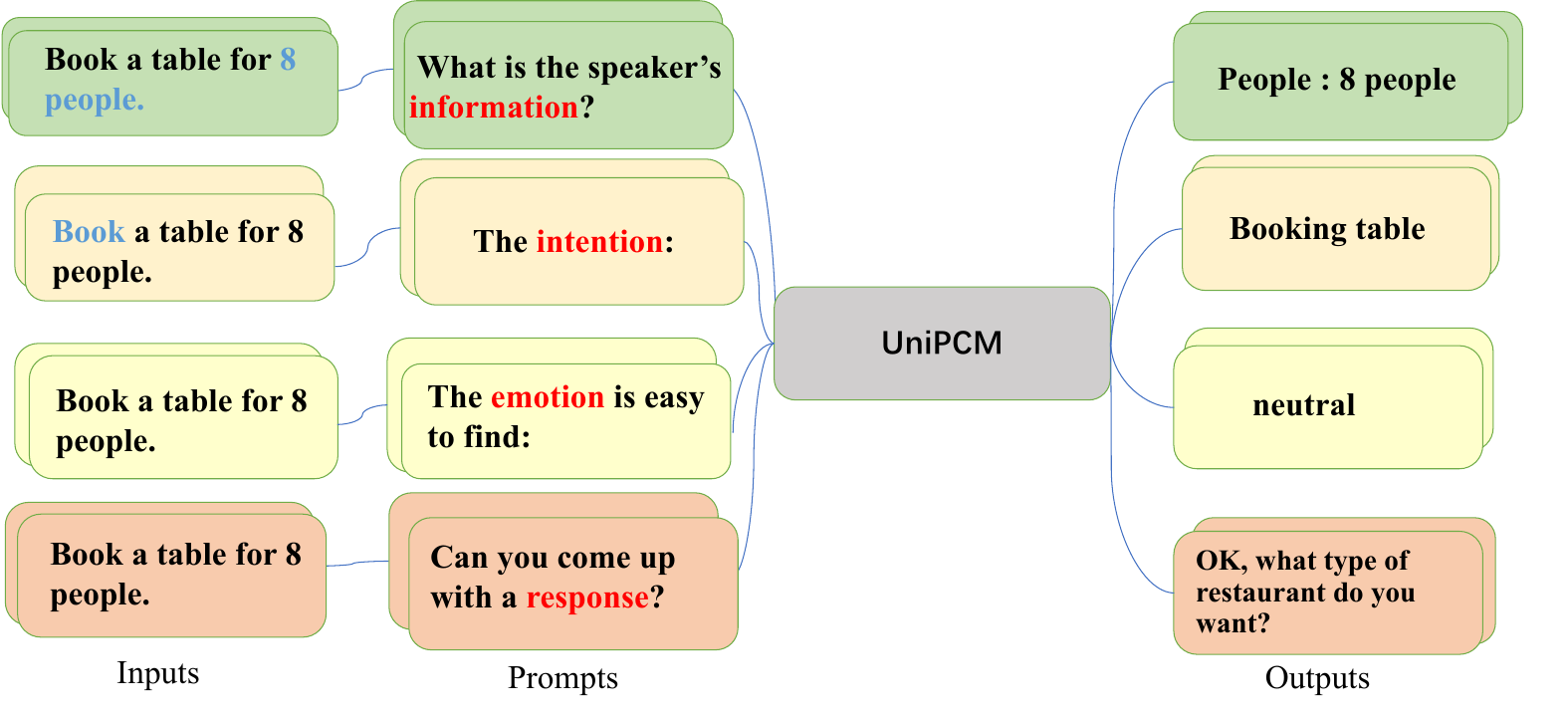}
	\caption{An illustration of our proposed model \modelname{}, which unifies all tasks into an 'input-prompt-output' format. Prompts are crucial as they help the model understand the task it should perform. }
	\label{fig:model}
\end{figure*}

\section{Related Work}

\subsection{Multi-task Language Model Pre-training}

Recent researches have shown that multi-task language model pre-training or pre-finetuning can greatly improve the model's transfer ability, resulting improved performance in few-shot and zero-shot settings \citep{raffel2020t5,wei2021finetuned,sanh2022multitask}. 
Although negative transfer may occur when the number of tasks is limited, the model will still benefit from pre-training if scaling the number of task \citep{aribandi2021ext5}.

To implement multi-task pre-training, some signals are given to the model to distinguish one task from another. Initially, researchers do multi-task pre-training using a unified text to text format directly \citep{raffel2020t5,lu-etal-2022-unified}. 
Simply adding the name of the task will help the model better understand the relation between task and reduce negative transfer problem \citep{zhang2022taskcompass}. Recent works use crowdsourcing prompts and instructions to perform multi-task pre-training, which achieves great success \citep{sanh2022multitask,wang2022super,ouyang2022training}. The resulting models show strong transfer ability, and can even chat with humans fluently in open domain \citep{ouyang2022training}.

Our work improves over the previous works in that we use automatically generated prompts instead of the crowdsourcing ones to enable multi-tasking, which reduces human labor as well as improves the quality of the prompts. Furthermore, we propose and formulate multi-prompt training mechanism, which relieves several problems in multi-task pre-training, including task imbalance, uneven data quality and difference between the importance of tasks. Moreover, we prove that multi-prompt training can improve model's performance on unseen prompts. 
\subsection{Automatic Prompt Generation}

It has been shown that prompt engineering can be of great benefit to reduce the gap between language model pre-training and finetuning on downstream tasks \citep{gao2021making,zhong-etal-2021-factual}. To reduce 
 human labor in prompt engineering, various approaches have been proposed to generate prompts automatically. AutoPrompt \citep{shin2020autoprompt} use gradient-based prompt search to automatically generates prompts. However, the prompt generated are not coherent, and may confuse models in multi-task scenarios.
Researchers proposed in \citep{gao2021making,zhou2022large} to use T5 \citep{raffel2020t5} or large language model to fill in the blank bewteent the input and output for automatically generating coherent prompts. However, the prompts generated do not necessarily contain task information and may be highly related to certain input case or dataset. 
Different from previous works, our work aims to generating prompts for multi-task pre-training. Therefore, our method \methodname{} models task in automatic prompt generation to help the model understand the relation between the  tasks and the prompts, as well as improve the quality of generated prompts. 


\subsection{Pre-training for Dialog Systems}
It has been shown that pre-training can greatly improve performance for dialog systems, improving coherency of the generated response and transfer ability \citep{roller2021recipes,zhang2020dialogpt,su-etal-2022-multi}. Models trained on large-scale online open-domain dialogues, for example, BlenderBot \citep{roller2021recipes}, DialoGPT \citep{zhang2020dialogpt} and Meena \citep{adiwardana2020towards}, can perform well on the chit-chat task, while models pre-trained on certain tasks can improve performance on corresponding tasks. For example, in task-oriented dialog, works like TOD-BERT \citep{wu2020tod}, CONV-BERT \citep{Mehri2020DialoGLUE}, PPTOD \citep{su-etal-2022-multi}, GALAXY \citep{he2021galaxy} improve the performance on relevant datasets. 

However to interact with human fluently in open-domain \citep{ouyang2022training}, the dialog system should not only be capable of various tasks, but also be robust to different input prompts. Recent progress in building powerful open-domain dialog systems mainly used crowdsourcing to annotate large-scale, multi-task datasets to improve the systems' performance
\citep{shuster2022blenderbot,ouyang2022training}. 
Different from their approaches, we propose to leverage the existing large scale datasets that are dialog-related to perform multi-task pre-training. \citet{chen2022dialogzoo} also trains their dialog foundation model over large scale dialog-related datasets. However, they do not aim to building a dialog system, therefore they do not improve their model's robustness to input prompts, neither do they evaluate their model's transfer ability in few-shot or zero-shot scenarios. In contrast, we use generated prompts to perform multi-task prompt pre-training to improve the model's transfer ability and robustness to different input prompts.

\subsection{Exploit Prompts for Low Resource Setting}%
Prompts can reduce the gap between language model pre-training and finetuning, therefore improving model's performance in downstream tasks, especially in few-shot and zero-shot settings
 \citep{gao2021making,cui-etal-2021-template,chen2022knowprompt}. Apart from that, pattern exploit training (PET), a self-training method leveraging multiple prompts, can greatly improve model's performance in low resource setting by perform semi-supervision training Different prompts can be used as different view for the case, and models finetuned with different prompt are used to ensemble pseudo labels on unlabeled data 
  \citep{schick2021exploiting}.
There are a few works that improve over the original pattern exploit training: \citet{schick2021is} extends PET to deal with labels that have multiple tokens, while \citet{tam2021improving} proposes to provide more
supervision and learn without task-specific unlabeled data. Our \pet{} contributes in reformulating PET to apply it to generative language model. Moreover, we combine PET with our multi-task prompt pre-trained model and applied multi-prompt training in the finetuning stage of PET, improving the accuracy of the generated pseudo labels.

\section{Method}

To pre-train our \modelname{}, we first 
unify all the dialog-related tasks into an 'input-prompt-output' format, which is shown in Figure \ref{fig:model}. Then we propose task-aware automatic prompt generation \methodname{} to generate high quality prompts for the pre-training.
Finally, based on the prompts and corpus, we pre-train our \modelname{} using the proposed multi-prompt training mechanism.

 





\subsection{Task-aware Automatic Prompt Generation}

 \begin{figure*}[t]
\centering
	\includegraphics[width=0.98\linewidth]{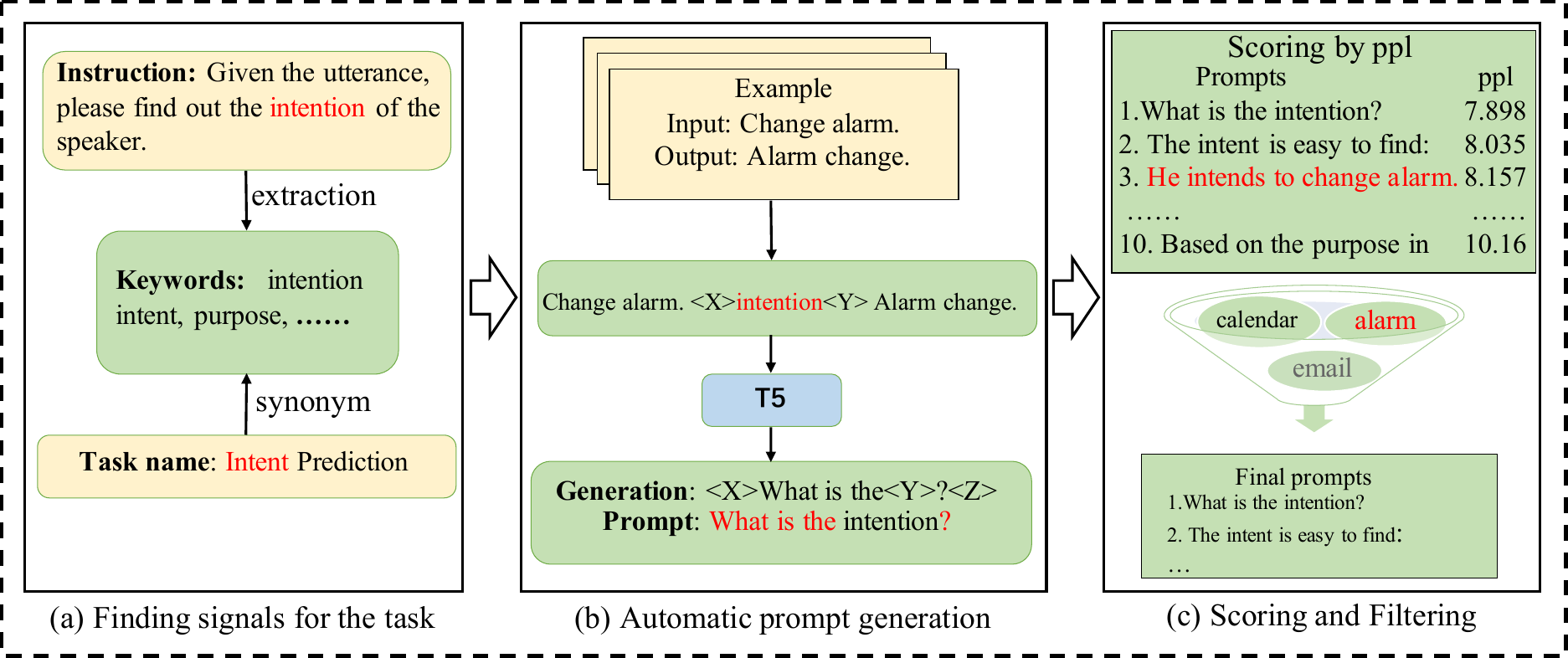}
	\caption{An illustration of our proposed method \methodname{}. (a) We collect existing signals to generate keywords, either by extracting from instructions or searching synonyms of the task name. (b) We automatically generate prompts based on input examples and task-related keywords using a T5 model, harvesting prompts by concatenating generated words after the sentinel token. (c)We select the prompts by average perplexity on task-related examples and filter out those prompts that may contain information about the labels.}
	\label{fig:main}
\end{figure*}


Our task-aware automatic prompt generation method \methodname{}, as illustrated in Figure \ref{fig:main}, mainly contains the following 3 parts:
\subsubsection{Finding Signals for Task Information}

Before generating prompts, we extract task-related signals to help us to find the information about the task $t$. In this work, we mainly focus on 3 kinds of signals that can be used as hint of the task for the model to generate prompts upon. We discuss their availability, limitation and effectiveness to generate prompts. 

\textbf{Instructions:}
Task descriptions, or instructions are usually available for datasets. Moreover, huge amounts of instructions are annotated or collected 
 by researchers or crowdsourcing workers \citep{wang2022super,ouyang2022training}.
 Instructions are usually long and difficult for language model to understand directly, therefore it's hard to directly use them as input to generate prompts in an unsupervised way. However, instructions contain almost all important information for the task and it's not hard to extract key information from it.

 In our work, we use tf-idf methods 
 to filter out irrelevant words. \footnote{We inplemented tf-idf using the gensim package: \hyperlink{gensim}{https://radimrehurek.com/gensim/}} 
 Then we get all the 1-gram, 2-grams and 3-grams of the remaining words to get scored by a Bert model \citep{devlin2019bert} using their similarity with the task name.
 The words that have similarity score above a threshold 
 are deemed as keyword and used to generate prompts. 

\textbf{Task Name:}
Task name is always available for task and very concise, ideal for automatic prompt generation. However, one task can only have one task name, making it difficult to generate diverse prompts. Therefore, we propose to use the thesaurus tool to paraphrase the task name to form diverse key words. Also, the task name is used to select the keywords extracted from the instruction, which has already been discussed.

\textbf{Keywords:}
Keywords are ideal input for automatic prompt generation, as keywords are both concise and representitive of the task information.
However, keywords are not readily available and should be inferred from other task-related information like instructions or the task name. If the quality or the number of the keywords generated by the instructions or the task name do not meet the researchers need,
researchers can quickly summarize the task and write some high-quality keywords themselves.

\subsubsection{Automatic Prompt Generation Using Keywords}

Getting the task signals (in the form of keywords in this work), we can generate prompts automatically using a pre-trained language model T5 \citep{raffel2020t5}. 
T5 is pre-trained to fill missing spans for a sentence. For example, given the input “Thank you <X> me to your party <Y> week”, T5 is trained to generate
“<X> for inviting <Y> last <Z>”, meaning that “for
inviting” replaces the placeholder <X> and “last”  replaces the placeholder <Y>. This is well suited for prompt generation, as we want to generate a prompt with the keywords that is coherent given input and output.

Given an instance of input-output pair $(X_{t},Y_{t})$ in task $t$, along with one of the keywords $k_{t}^{i}$, we define a transform $\mathcal{T}(X_{t},Y_{t},k_{t}^{i})$
\footnote{In fact, multiple reasonable transforms can be defined. In our experiments, we use the transform $\Tilde{\mathcal{T}}(X_{t},Y_{t},k_{t}^{i}) = X_{t}, Y_{t}, k_{t}^{i} \to \langle X\rangle X_{t} \langle Y\rangle k_{t}^{i} \langle Z\rangle Y_{t}$, along with the transform $T$ introduced in the paper to better model tasks that need a prefix like question answering.}:

\begin{align} 
& X_{t}, Y_{t}, k_{t}^{i} \to X_{t} \langle X\rangle k_{t}^{i} \langle Y\rangle Y_{t} 
\label{eq:input-format}
\end{align}
where $\langle X \rangle,\langle Y \rangle$ 
are sentinel tokens for T5 generation.  We generate the prompts according to the T5 generation probability $P_{T5}(\mathcal{T}(X_{t},Y_{t},k_{t}^{i}))$ and harvest the prompts generated after the sentinel tokens. The final prompts are reorganized as $ P_{t} = x \oplus k_{t}^{i} \oplus y $, where $\oplus$ denotes the concatenation of token sequence and $x,y$ are the corresponding content generated after sentinel tokens $\langle X \rangle,\langle Y \rangle$ by the T5 model.

For one single input-output instance and one keyword, we get the top 5 prompts according to generation probability. Using multiple instances and multiple keywords, we generate numerous prompts for selection. To avoid generating prompts that are specific to one single input instance and overlook the task information, we only retain the prompts that appear multiple times, which is empirically set as 2 in our work, in different instances.

\subsubsection{Scoring and Filtering of Generated Prompts}

As it is discussed in section \ref{sec:model}, given input $X_{t} $ and prompt $P_{t} $ in task $t$, the probability of generating the correct answer $Y_{t}$: $p(Y_{t}|X_{t},P_{t})$ should be optimized, therefore we evaluate the quality of the generated prompts according to the probability, as all of $Y_{t},X_{t},P_{t}$ are available and the probability can be directly calculated.
We choose those prompts $P_{t}$ that has higher average log probability $\sum\log(p(Y_{t}|X_{t},P_{t}))$ among the datasets in task.

After that we filter out prompts that may contain certain biased information about the output $Y_{t}$ using a prohibited words list extracted from the outputs. The prohibited words mainly fall into the classification type class, as the output $Y_{t}$ is selected from certain labels. 
For example, in emotion classification task, the word "positive" or "negative" is often generated by the model, as the output $Y_{t}$ contains those two words with high frequency. However, using such prompts for output generation, the result will be biased, reducing the generated performance. Therefore, we filter out those case-sensitive prompts to encourage those prompts that accurately reflect the task.

\subsection{Multi-task Prompt Pre-training}

\label{sec:multitask}

Using the generated prompts, as well as the collected corpus, we perform multi-task prompt pre-training. 
With the training instances $(X_{t}^{i},Y_{t}^{i})(i=1,2, \cdots, N_{t})$ from $K$ different tasks and the generated prompts $P_{t}^{j}(j=1,2, \cdots, M_{t})$, the objective function of the pre-training can be written as:

\begin{align} 
\mathcal{J}_{\theta} = \sum \limits_{t=1}^{K} \sum \limits_{i=1}^{N_{t}} \sum \limits_{j=1}^{M_{t}} \log p(Y_{t}^{i}|X_{t}^{i},P_{t}^{j})
\label{eq:training-objective}
\end{align}



Note that in Eq. \ref{eq:training-objective}, we propose to use multi-prompt training, which means applying multiple prompts to one single input instance:
$\sum \limits_{j=1}^{M_{t}} \log p(Y_{t}^{i}|X_{t}^{i},P_{t}^{j})$.  
The benefits of which is discussed in sec \ref{multi-prompt-benefit}. 
 
However it's not necessary to apply all the $M_{t}$ prompts available to one single case, as the prompts $P_{t}^{j}(j=1,2, \cdots, M_{t})$ are representation of the task $t$ and have similar embeddings in the latent task space. Therefore a subset of ${P_{t}^{j}}$ can be randomly sampled, resulting in ${\Tilde{P}_{t}^{j}}(j=1,2, \cdots, \Tilde{M_{t}})$. The loss $\sum \limits_{j=1}^{M_{t}} \log p(Y_{t}^{i}|X_{t}^{i},P_{t}^{j})$ can be approximated by $\frac{M_{t}}{\Tilde{M_{t}}}\sum \limits_{j=1}^{\Tilde{M_{t}}} \log p(Y_{t}^{i}|X_{t}^{i},\Tilde{P}_{t}^{j})$ to save calculation time. 

If the ratio $\frac{M_{t}}{\Tilde{M_{t}}}$ is not added, we can simply adjusting the weights of datasets or tasks in pre-training by adjusting the number of prompts applied. It is beneficial as some tasks or datasets are deemed more important by the researchers. Adding more prompts to those tasks or datasets can make the model better focus on them.



\subsection{Prompts for Semi-Supervised Training: \pet{}}
\label{sec:pet}

To utilize numerous and diverse generated prompts, as well as the pre-trained model that performs well on those prompts, we perform \pet{} \citep{schick2021exploiting} for semi-supervised training. We adapt the origin PET method to better utilize multiple prompts, as well as fitting our pretrained model.

For the generated prompts $P={{P^{j}}(j=1,2, \cdots, M)}$, we use a partition of $P$, ${P_{1}}, {P_{2}}, \cdots ,{P_{k}}$ to train $k$ voting models for ensembling. The $l$th voting model $M_{l}$ are finetuned from the pre-trained model on the annotated part of data $(X^{i},Y^{i})(i=1,2, \cdots, N_{a})$ with the prompt sets $P_{l}$, the loss function as follows:

\begin{align} 
\mathcal{J}_{\theta}^{l} = \sum \limits_{i=1}^{N_{a}} \sum \limits_{j=1}^{|P_{l}|} \log p_{M_{l}}(Y^{i}|X^{i},P_{l}^{j})
\label{eq:pet-finetuning-objective}
\end{align}

To generate pseudo labels on unannotated data, we ensemble the outputs generated by voting models given all input instances and prompts:

\begin{align} 
\Tilde{Y^{i}} = ensemble(\{\Tilde{Y^{i}_{j}}\}) ,\Tilde{Y^{i}_{j}}\sim p_{M_{l}}(\Tilde{Y^{i}}|X^{i},P_{l}^{j})
\label{eq:pet-sampling}
\end{align}

where we use majority voting method to perform ensembling for the labels generated. Sampling is used in (\ref{eq:pet-sampling}) to increase diversity of the generated label, helping us to distinguish those instances and labels that are deemed uncertain by the model. Because we finetune the voting models on a model pre-trained over all prompts and we use multi-prompt training to finetune the voting models in (\ref{eq:pet-training-objective}), the accuracy of the voting models is greatly improved, therefore advancing the quality of the pseudo labels generated.

The $N_{p}$ pseudo labels are used to train the model, along with the annotated data, to improve the model's performance:
\begin{align} 
\mathcal{J}_{\theta} & = \sum \limits_{i=1}^{N_{a}} \sum \limits_{j=1}^{M} \log p(Y^{i}|X^{i},P^{j}) \nonumber\\
&+ \sum \limits_{k=1}^{N_{p}} \sum \limits_{j=1}^{M} \log p(\Tilde{Y^{k}}|X^{k},P^{j})
\label{eq:pet-training-objective}
\end{align}

\section{Experiments}

\subsection{Baseline\&Benchmark}
To comprehensively evaluate our \modelname{}, we carefully choose ten downstream datasets in six tasks, mainly evaluating the model's ability in dialog understanding, response generation, and comprehensive ability. 

\subsubsection{Dialog understanding} 
Dialog understanding is crucial for building a high-quality dialog system as it's impossible to generate high-quality responses without having a good understanding of the context. DialoGLUE \citep{Mehri2020DialoGLUE} is a benchmark that comprehensively evaluates the dialogue understanding ability of a dialog system, which consists of four tasks: slot filling
(REST8K \citep{coope2020span}, DSTC8 \citep{rastogi2020towards}), intent prediction (BANKING77 \citep{casanueva2020efficient}, CLINC150 \citep{larson2019evaluation}, HWU64 \citep{liu2021benchmarking} ), semantic parsing (TOP \citep{gupta2018semantic}), and dialog state tracking (MultiWOZ2.1 \citep{eric2020multiwoz}). We follow the original preprocessing and evaluating scripts of \citet{Mehri2020DialoGLUE}, except that we modify the implementation to a sequence-to-sequence generation format to fit the model's pre-training. The evaluation metrics for slot filling, intent prediction and semantic parsing are F1, accuracy and exact match respectively. For dialog state tracking task of Multiwoz, we apply our model to the SOTA generative baseline SDP-DST \citep{lee-etal-2021-dialogue} and joint goal accuracy (JGA) is reported.
 Apart from T5 \citep{raffel2020t5} (we pre-trained our model upon a T5-base model), we choose SPACE-2 \citep{he-etal-2022-space} and 
 Flan-T5 \citep{chung2022scaling} as our baselines, as SPACE-2 represent the SOTA pre-trained results targeting task-oriented dialog understanding, while Flan-T5 is a general-purpose pre-trained language model using instruction-tuning. The results of TOD-BERT \citep{wu2020tod} and the best variant of ConvBERT \cite{Mehri2020DialoGLUE} are also reported for comparison.

\subsubsection{Response Generation}
Open-domain response generation, or chit-chat, is also an important skill for building a high-quality dialog system. 
We evaluate our model on two classic chit-chat datasets PersonaChat \citep{zhang2018personalizing} and DailyDialog \citep{li2017dailydialog}. We follow the preprocessing and evaluation scripts of FSB \citep{madotto2021few}, BLEU \citep{papineni2002bleu}, word-level F1 and Rouge-L \citep{lin2004rouge} reported. 
We choose DialoGPT \citep{zhang2020dialogpt} and PPTOD \citep{su-etal-2022-multi} as our baseline.
\subsubsection{Comprehensive ability} We evaluate the comprehensive ability of a dialog system on the Multiwoz end to end generation task (End2End) \citep{budzianowski2018large}. In End2End task, the model needs to track the use's state, understand user's intention, decide the best responding strategies and generate coherent response, which is quite challenging. Multiple dialog skills, such as intent prediction, dialog state tracking, policy optimization, and response generation, are necessary to complete the task.
 We apply our model to the SOTA method MTTOD \citep{lee-2021-improving-end} and use the official evaluation scripts \footnote{\hyperlink{multiwoz}{https://github.com/budzianowski/multiwoz}} given by \citep{nekvinda2021shades}. We compare our results to LABES \citep{zhang-etal-2020-probabilistic}, SOLOIST \citep{peng-etal-2021-soloist}, UBAR \citep{yang2021ubar} and PPTOD \citep{su-etal-2022-multi}.


	\begin{table*}[t]
  \caption{Results of seven datasets from the DialoGLUE benchmark in low-resource and full data setting. $^{*}$ denotes the model is specified for understanding task in TOD only. $^{\dagger}$ denotes we fix a bug in the original scripts, resulting higher score in DSTC8 dataset and we exclude the dataset in the avg score for fair comparison. }
		\centering
		\resizebox{1.0\linewidth}{!}{
				\begin{tabular}{cccccccccc}
					\toprule
				Setting	&Model & avg & BANKING77 &HWU64 &CLINIC150 &REST8K &DSTC8 $^{\dagger}$ &TOP &MULTIWOZ  \\
					\midrule
			\multirow{6}{*}{10-shot data}&T5  & 76.52 &76.01 &81.77 &88.36 &85.31 &74.72 &76.03 &51.63 \\
   
      &TOD-BERT $^{*}$ &79.96 &85.99& 86.74& 93.07& 87.62 &50.19& 77.77 &48.54\\
      &ConvBERT $^{*}$&78.72 &85.06   & 85.69 &93.06 &87.58 &44.36 &72.01 & 48.89\\
      &SPACE-2 $^{*}$ &\underline{81.91} &88.31   & \underline{88.85} &\underline{95.22} &88.85 &54.41 &\underline{79.55} & 50.70\\
      &Flan-T5 & 80.68 & 84.48   & 86.88 &91.80 &\underline{90.59} &\underline{78.68} &76.78 & \underline{53.52}\\      
        & \textbf{\modelname{}} 
        &\textbf{83.66} &\textbf{90.16}   & \textbf{90.05}   &\textbf{95.78} &\textbf{92.62} &\textbf{83.27} &\textbf{79.63} & \textbf{53.73} \\
					\midrule
                        \midrule
                    \multirow{6}{*}{Full data}&T5 & 85.70 &92.60 &91.07 &96.49 &95.95 &93.60 &81.41 &56.66 \\
 &TOD-BERT$^{*}$ & 85.43 & 93.02 &89.87 &95.93 &95.53 &90.05& 81.90 &56.30\\
 &ConvBERT$^{*}$ &86.17 &93.44 &92.38 &97.11 &95.44 &91.20 &82.08 &56.56\\
					 &SPACE-2$^{*}$ &87.56 &\textbf{94.77}   & \textbf{94.33} &\textbf{97.80} & 96.20 &91.38 &82.74 & \textbf{59.51}\\
	               &Flan-T5&86.99 &93.47   & 92.37 &96.71 &\underline{96.41} &\underline{94.51} &\underline{84.32} & 58.68\\  				
      & \textbf{\modelname{}}&\textbf{87.59} &\underline{94.41}  &\underline{93.40} &\underline{97.47}&\textbf{96.92}& \textbf{96.15} &\textbf{84.58} & \underline{58.76}\\
					\bottomrule
			\end{tabular}}
			\label{dialoglue-results}
	\end{table*}

\begin{table}[t]
 \caption{Full data and few-shot results on Multiwoz2.0 End2End task, inform, success, BLEU and combined score 
 are reported. }
		\centering
		\resizebox{\linewidth}{!}{
			\begin{tabular}{cccccc}
				\toprule
                &\multicolumn{4}{c}{MultiWOZ2.0 End2End }\\
                \midrule
	Setting	&Model&Inform &Success &BLEU &Combined score \\
				\midrule
   \multirow{6}{*}{Full data} & LABES &68.5  &58.1 &18.9 &82.2 \\
&SOLOIST &82.3  &72.4 &13.6 &90.9 \\
&UBAR &83.4  &70.3 &17.6 &94.4 \\
&PPTOD &83.1	&72.7 &18.2 &96.1 \\
&MTTOD &85.9	&76.5  &19.0	 &100.2 \\
&\modelname{} (ours) &\textbf{88.3}  &  \textbf{76.8}&\textbf{19.2} &\textbf{101.8} \\
\midrule
\midrule
\multirow{2}{*}{Few shot(10\%)} &MTTOD &66.8  &52.8 &\textbf{15.7} &75.5 \\
&\modelname{} (ours) &\textbf{68.4} &\textbf{57.2} &14.9 &\textbf{77.7} \\

         \bottomrule	
	\end{tabular}}
	\label{multiwoz result}
	\end{table}
 
 \begin{table}[t]
 \caption{ Few-shot and Zero-shot results on Personachat and DailyDialog dataset (task: chit-chat). BLEU, word-level F1 
 and Rouge-L are reported. }
		\centering
		\resizebox{\linewidth}{!}{
			\begin{tabular}{cc ccc ccc}
				\toprule
                \multicolumn{2}{c}{Model Configuration} &\multicolumn{3}{c}{Persona}&\multicolumn{3}{c}{DailyDialog}\\
                 \cmidrule(lr){1-2} \cmidrule(lr){3-5} \cmidrule(lr){6-8}
	Setting	&Model& BLEU &F1 &Rouge-L &BLEU &F1 &Rouge-L \\
				\midrule
   \multirow{4}{*}{Zero-shot} & T5 (baseline) &0.94  &15.24 &9.16 &0.29  &9.76 &8.51 \\
&PPTOD &0.70  &13.83 &10.74 &0.39	&10.44 &10.14 \\
&DialoGPT &0.57	&9.61 &11.83 &0.45	&15.18 &18.99 \\
&\modelname{} (ours) &\textbf{1.15}	&\textbf{16.45}  &\textbf{18.25}	 &\textbf{0.85}	&\textbf{17.81}  &\textbf{21.04} \\
\midrule
\midrule
\multirow{4}{*}{Few-shot(10\%)} & T5 (baseline) &1.76  &17.18 &18.14 &0.53  &12.62 &16.44 \\
&PPTOD &1.85  &17.44 &17.75 &0.39  &14.58 &17.65 \\
&DialoGPT &1.23	&14.74 &18.39 &0.77	&16.35 &18.16 \\
&\modelname{} (ours) &\textbf{2.41}	&\textbf{19.16}  &\textbf{18.81}	 &\textbf{0.81}	&\textbf{18.04}  &\textbf{21.23} \\

         \bottomrule	
	\end{tabular}}
		\label{chitchat results}
	\end{table}

 \subsection{Implementation}
\label{sec:implementation}


\begin{table*}[t]
\caption{ Statistics of tasks, datasets, and prompts in \dataname{}.}
		\centering
		\resizebox{0.99\linewidth}{!}{
			\begin{tabular}{cc cccccccccc}
				\toprule
				Task type &Intent   &Dialog state tracking &Emotion  &Summary  &Question answering &Generation &Response &Multiple choice &Text2sql & Grounded dialog& Total\\
    \midrule
    Tasks &Intent  &DST, slot filling &Emotion &Summary  &DialQA, DocQA &Generation &Response, Chat &Multiple choice &Text2sql & TOD, US, KG-dial &15 \\
	Number of prompts& 37 & 33 & 14 & 11 & 35 & 51 & 27 & 39 &29 &27 & 303\\		
    Number of datasets& 22 & 21 & 7 & 5 & 12 & 4 &23 & 3 & 2 &23 & 122\\
  Number of instances 
  & 1,382,413 & 4,382,314 & 171,353 & 449,995 & 460,681 & 198,999 & 16,555,894 & 44,992 &19,059 & 2,959,786 & 26,625,486\\
         \bottomrule	
		\end{tabular}}
		\label{tab:statistics}
	\end{table*}
 
\subsubsection{Building pre-training corpus}

  To perform multi-task 
 pre-training for conversation model, we collect \dataname{} \footnote{
We collect datasets from \hyperlink{datasets}{https://huggingface.co/datasets}, \hyperlink{parlai}{https://www.parl.ai/docs/tasks.html} and GitHub repositories on  \hyperlink{github}{https://github.com/}.}, which contains 122 dialog-related from 15 dialog-related tasks. The tasks in \dataname{} mainly fall into three categories: task-oriented dialog related (intent prediction, dialog state tracking and grounded dialog), open-domain chit-chat, and other dialog-related datasets. 

 Task-oriented dialog is extensively studied by previous researchers, resulting in abundant annotated datasets. We make full use of the annotated information as we leverage prompts to convert a turn in a dialog into multiple training instances, as shown in Figure \ref{fig:model}.

 Open-domain chit-chat datasets are important for improving the generation ability of pre-trained conversation models. We use the datasets collected in \citet{he2022unified} \footnote{\hyperlink{space3}{https://github.com/AlibabaResearch/DAMO-ConvAI/tree/main/space-3}} as the datasets are competitive in quality and quantity. However, instead of viewing those datasets as unannotated data for semi-supervised training for task-oriented dialog, we train the response generation task on those datasets, leveraging the coherency of open-domain chat datasets. 
 
 To extend the model's ability, we collect other datasets that can improve the model's skills. 
 Emotion classification, summary, natural language generation, and text2sql are important skills for dialog systems in real-life scenarios,
 while question answering and multiple choice have similar format as dialog and will yield positive transfer in co-training \cite{aribandi2021ext5}.

The statistics of the tasks and datasets, as well as the generated prompts, are shown in Table \ref{tab:statistics} and the details of the tasks and datasets can be found in Table \ref{pretrain-datasets}.

\subsubsection{Pre-training}
\label{sec:dataset}


We pre-train our conversation model \modelname{} on the collected corpus \dataname{}, the details of which shown in Table \ref{pretrain-datasets}. The maximum sequence length of input context is set to 256. The batch size is set to 64 and an AdamW optimizer is employed for optimization with a constant learning rate of 2e-5. The pre-training is performed on eight 80GB Tesla-A100 GPU and takes about 72 hours.

\begin{table*}[t]
  \caption{Pre-training tasks and datasets in \dataname{}.}
		\centering
		\resizebox{\linewidth}{!}{
			\begin{tabular}{c|l}
				\toprule
				\textbf{Task} &\textbf{ Datasets}\\
				\midrule
				Natural language generation 
    & web\_nlg \cite{ferreira20webnlg}, dart \cite{nan-etal-2021-dart}, e2e\_nlg \cite{dusek2020e2e}, common\_gen \cite{lin-etal-2020-commongen}\\
    \midrule
    Summary 
    & dialogsum \cite{chen-etal-2021-dialogsum}, xlsum \cite{hasan-etal-2021-xl}, xwikis \cite{clads-emnlp}, wiki\_lingua \cite{ladhak-etal-2020-wikilingua}, Samsum \cite{gliwa2019samsum} \\
    \midrule
    Slot filling 
    &Restaurant8k \cite{coope2020span}, TOP \cite{gupta2018semantic}, DSTC8 \cite{rastogi2020towards}, ATIS \cite{hemphill1990atis}, CrossNer \cite{liu2021crossner}, FB\_TOD\_SF \cite{schuster2019cross}, 
    \\&MIT-movies-eng \cite{liu2013asgard}, MIT-movies-eng \cite{liu2013asgard}, MIT-movies-trival10k \cite{liu2013asgard}, MIT-restaurant \cite{liu2013asgard}, SNIPS \cite{coucke2018snips},
    \\
    \midrule
    Intent prediction 
    &BANKING77 \citep{casanueva2020efficient}, CLINC150 \citep{larson2019evaluation}, HWU64 \citep{liu2021benchmarking}, FB\_TOD\_SF \cite{schuster2019cross}, SNIPS \cite{coucke2018snips}, TOP \cite{gupta2018semantic}, 
    \\
   &MultiWOZ2.2 \cite{zang2020multiwoz}, SGD\cite{rastogi2020towards}, WOZ \cite{mrkvsic2017neural}, SimJoint \cite{shah2018bootstrapping}, MultiWOZ\_synthesis \cite{campagna2020zero}, SwDA \cite{stolcke2000dialogue},\\
   &DailyDialog \cite{li2017dailydialog}, DSTC2 \cite{williams2016dialog}, DSTC3 \cite{williams2016dialog}, InCar \cite{eric2017key} ,PersuaGOOD \cite{wang2019persuasion},  Frames \cite{el2017frames},\\
   &MulDoGo \cite{peskov2019multi}, BiTOD \cite{lin2021bitod}, MSRe2e \cite{li2018microsoft},\\
   \midrule
   Dialog state tracking 
   &SGD \cite{rastogi2020towards}, TaskMaster1 \cite{byrne2019taskmaster},  TaskMaster2 \cite{byrne2019taskmaster}, TaskMaster3 \cite{byrne2019taskmaster}, WOZ \cite{mrkvsic2017neural}, KETOD \cite{chen2022ketod},\\
   &MulDoGo \cite{peskov2019multi},
InCar \cite{eric2017key}, SimJoint \cite{shah2018bootstrapping},MultiWOZ2.2 \cite{zang2020multiwoz},
    \\
    \midrule
    Multiple choice 
   & Commensense-qa \cite{talmor-etal-2019-commonsenseqa}, Cosmosqa \cite{huang-etal-2019-cosmos}, Meld \cite{poria2019meld}\\
   \midrule
   Emotion classification 
   &DailyDialog \cite{li2017dailydialog}, Go-emotion \cite{demszky2020goemotions}, Meld \cite{poria2019meld}, SentiHood \cite{saeidi2016sentihood}, MAMS \cite{jiang-etal-2019-challenge}, ASTE \cite{xu-etal-2021-learning}, \\
   &RECCON \cite{poria2021recognizing}\\
   \midrule
   Document-based question answering 
   &SQuAD \cite{rajpurkar2016squad}, QuAC \cite{choi-etal-2018-quac}, NarrativeQA \cite{narrativeqa}, Race \cite{lai-etal-2017-race}\\
   \midrule
   Dialog-related question answering 
   &DREAM \cite{sun2019dream}, Molweni \cite{li2020molweni}, DialogRE \cite{yu2020dialogue}, FriendsQA \cite{yang2019friendsqa}, DDRel \cite{jia2021ddrel}, ReadingComprehension \cite{ma2018challenging},\\
   &RECCON \cite{poria2021recognizing}, WizInt \cite{komeili-etal-2022-internet}\\
   \midrule
   Chit-chat \&
   & Mutual \cite{cui2020mutual}, ABCD \cite{chen2021action}, AirDialog \cite{wei2018airdialogue}, CCPE \cite{radlinski2019coached}, MetalWOZ \cite{shalyminov2020fast}, CMU\_DoG \cite{zhou2018dataset}, \\
   Response generation 
   & CoQA \cite{reddy2019coqa}, CoSQL \cite{yu2019cosql}, doc2dial \cite{feng2020doc2dial}, DSTC10-track2 \cite{kim2021robust}, DSTC10-track3 \cite{kottur2021simmc}, MedicalDialog \cite{zeng2020meddialog}, \\
   &Self-Dialog \cite{fainberg2018talking}, WOW \cite{dinan2018wizard}, TopicChat \cite{gopalakrishnan2019topical}, Persona-Chat \cite{zhang2018personalizing}, MulDoGo\_un \cite{peskov2019multi}, \\
   &CSQA \cite{saha2018complex}, AmazonQA \cite{ijcai2019amazonqa}, ChitChat \cite{will2020conversational}, EmpatheticDialog \cite{rashkin-etal-2019-towards}, 
   CommonsenseDialog \cite{zhou2021commonsense},\\
   &ConvQuestions \cite{kacupaj2021conversational}, MMD \cite{saha2018towards} \\
   \midrule
   Knowledge-grounded dialog 
   & Soccer-kgdial \cite{chaudhuri2019using}, Incar-kgdial \cite{chaudhuri2019using}, WizInt \cite{komeili-etal-2022-internet}, KETOD \cite{chen2022ketod} \\
   \midrule
   Text to SQL 
   & Spider \cite{yu2018spider}, Sparc \cite{yu2019sparc} \\
   \midrule
   Task oriented dialog \& 
   & TaskMaster1 \cite{byrne2019taskmaster},  TaskMaster2 \cite{byrne2019taskmaster}, TaskMaster3 \cite{byrne2019taskmaster}, 
   SwDA \cite{stolcke2000dialogue}, FusedChat\cite{young2022fusing}, Frames \cite{el2017frames}, \\
   User simulation 
   & MultiWOZ2.2 \cite{zang2020multiwoz}, SGD\cite{rastogi2020towards}, WOZ \cite{mrkvsic2017neural}, SimJoint \cite{shah2018bootstrapping}, MultiWOZ\_synthesis \cite{campagna2020zero}, MulDoGo \cite{peskov2019multi},\\
   &DailyDialog \cite{li2017dailydialog}, DSTC2 \cite{williams2016dialog}, DSTC3 \cite{williams2016dialog}, InCar \cite{eric2017key} ,PersuaGOOD \cite{wang2019persuasion}, BiTOD \cite{lin2021bitod} \\
   &MSRe2e \cite{li2018microsoft},\\
    \bottomrule	
    \end{tabular}}
	\label{pretrain-datasets}
\end{table*}

\subsubsection{Downstream tasks}
For downstream tasks, we finetune \modelname{} following the corresponding baseline scripts. For each few-shot and zero-shot experiment, we exclude the training data other than the few-shot data in the pre-training datasets accordingly to avoid unfair data use. 
During testing, we test the model with 5 random prompts sampled from all the available prompts for each testinig instance (if the prompts are used). We view the results as 5 independent experiments and the mean result of the performance is reported as the final result. The variance of the experiment is reduced as we take the mean results of 5 experiments. Moreover, to achieve high score under this testing setting, the model needs to perform well on all the available prompts. The resulting high performance proves that our model is robust to input prompts.

 \subsection{Main Results}

We conduct our experiments on the baseline and benchmarks mentioned above.
The implementation detail is shown in Sec. \ref{sec:implementation}.

\subsubsection{DialoGLUE Results }
As shown in Table \ref{dialoglue-results}, our model \modelname{} excels at few-shot setting, improving \textbf{7.14\%} on average scores over the T5 baseline, achieving SOTA results on all 7 datasets of DialoGLUE and improve by \textbf{1.75\%} over the previous SOTA result SPACE-2 on average scores.


For the full data setting, our model is competitive, achieving the best average scores among the strong baselines 
and consistantly outperforms Flan-T5 on 
 all datasets, which demonstrate the efficacy of our pre-training methods. 
It is worth noticing that SPACE-2 performs quite well on this task, which is mainly because its TOD targeted modelling, which makes the model restricted to understanding task in 
 TOD datasets. 

\subsubsection{MultiWOZ2.0 End2End Results}
As shown in Table \ref{multiwoz result}, our model \modelname{} improves over the previous SOTA model MTTOD in both full data and few shot scenarios by 1.6 and 2.2 on combined score respectively. The model's improvements mainly fall in the Inform and Success, implying that the pre-training improves the model's dialog understanding and decision-making ability. Meanwhile, the few-shot improvements are not so remarkable as in DialoGLUE datasets, probably resulting from the delexicalization preprocessing used in MultiWOZ \citep{zhang-2020-labes}, making the language used in this dataset slightly different from those in other datasets in pre-training.

\subsubsection{Chit-chat Results}
As shown in Table \ref{chitchat results}, 
\modelname{} consistently improves over all of the baseline results in zero-shot and few-shot settings in Persona and DailyDialog. The results imply that combining open-domain chat datasets in the multi-task pre-training procedure will improve the model's ability to perform open-domain chatting. Meanwhile, the performance of PPTOD, a model trained on task-oriented dialog datasets only, does not improve over the T5 baseline on chit-chat tasks, which shows the importance of combining open-domain chit-chat tasks in pre-training.

\subsection{Analysis and Ablation Study}

\subsubsection{Ablation study for \modelname{} in few-shot setting }

It has been shown in Table \ref{dialoglue-results} that \modelname{} excels at few-shot setting, and we want to have a full understanding of why \modelname{} achieves great performance in few-shot setting.  We get three main conclusions from the ablation study shown in Table \ref{tab:ablation}: \textbf{(1) Using multi-prompt training in finetuning stage greatly helps the model's performance in few-shot setting},
achieving 2.98\% gain. \textbf{(2) Using multi-prompt training in pre-training stage will help the model learn better in multi-task scenario.} Although using one human-written prompt in the pretraining stage will help improve the dialog understanding ability by 1.20\%, by using multi-prompt training in the pre-training stage, 
the results improve by \textbf{3.08\%}, which shows that using multi-prompt training in the pre-training stage will greatly benefit the model's performance in downstream task.
\textbf{(3) PET (introduced in Sec. \ref{sec:pet}) will help in low-resource setting.}
Adding \pet{} improves by 1.08\% over the strong baseline, which shows that our generated prompts can help model better utilize unlabeled data by using \pet{}.

\begin{table}[t]
  \caption{Ablation study on six datasets from the DialoGLUE benchmark in low-resource setting (10-shot data), MP means multi-prompt training in the finetuning stage, PT means pre-training, MPPT means multi-prompt training in the pre-training stage. $^{*}$ denotes it is the \modelname{}. }
		\centering
		\resizebox{1.0\linewidth}{!}{
				\begin{tabular}{lcccccccc}
					\toprule
				Method & avg & BANKING77 &HWU64 &CLINIC150 &REST8K &DSTC8 &TOP\\ 
					\midrule
			T5 & 76.52 &76.01 &81.77 &88.36 &85.31 &74.72 &76.03 \\
      \textbf{+} MP & 79.50 &83.77   & 85.02 &91.67 &88.24 &79.67 &76.69 \\
      \textbf{+} MP \textbf{+} PT  & 80.70 &83.70   & 85.86 &92.73 &90.79 &80.41 &78.64 \\
         \textbf{+} MP \textbf{+} MPPT & 82.58 &87.92   & 88.74 &94.76  &91.55 &82.87 &78.75\\
         \textbf{+} MP \textbf{+} MPPT \textbf{+} PET$^{*}$ 
         &\textbf{83.66} &\textbf{90.16}   & \textbf{90.05}   &\textbf{95.78} &\textbf{92.62} &\textbf{83.27} &\textbf{79.63} \\
					\bottomrule
			\end{tabular}}
			\label{tab:ablation}
	\end{table}
\subsubsection{Finetuning with multiple prompts. }
Although we have shown in Table \ref{dialoglue-results} that multi-prompt training will greatly improve the model's performance of finetuning in few-shot setting, it is not clear how the number of prompts available will influence the final results. From Table \ref{promt-num-compare}, we can see that simply applying 1 prompt will increase by 2.306\% on test accuracy. Moreover, applying a small number of prompts (7) can greatly improve the test accuracy (4.643\%). To manually select prompts that are deemed better by human experts will not help much (0.323\%). Moreover, using a large number of prompts (25) will improve a little over fewer prompts result (0.811\%). Therefore in \pet{}, we propose to use subsets of prompts to finetune the voting models, which will yield the best performance.

\begin{table}[t]
  \caption{Few-shot(10\%) results on BANKING77 dataset using different numbers of prompts. For 1 prompt setting, we report the average scores of randomly selected prompt to reduce variance.}
		\centering
		\resizebox{\linewidth}{!}{
			\begin{tabular}{cccccc}
				\toprule
				Number of Prompts& 0  &1(avg) & 7(random) & 7(selected) &25\\
				\midrule
				Test Acc & 76.006 & 78.312& 82.955 &83.279 &\textbf{83.766} \\
         \bottomrule	
		\end{tabular}}
		\label{promt-num-compare}
	\end{table}
 \subsection{Automatically generated Prompts}

 Using the 494 keywords extracted from the Super-Instruction datasets \citep{wang2022super}, we generate 3423 prompts on 74 tasks. However, as our work mainly focus on pre-training a conversation model, we mainly evaluate the 303 prompts used in pre-training.
 The rest of generated prompts will be released with our codes and can be further studied.





\subsubsection{Visualization of Generated Prompts}
\begin{figure}[t]
\centering
\includegraphics[width=0.98\linewidth]{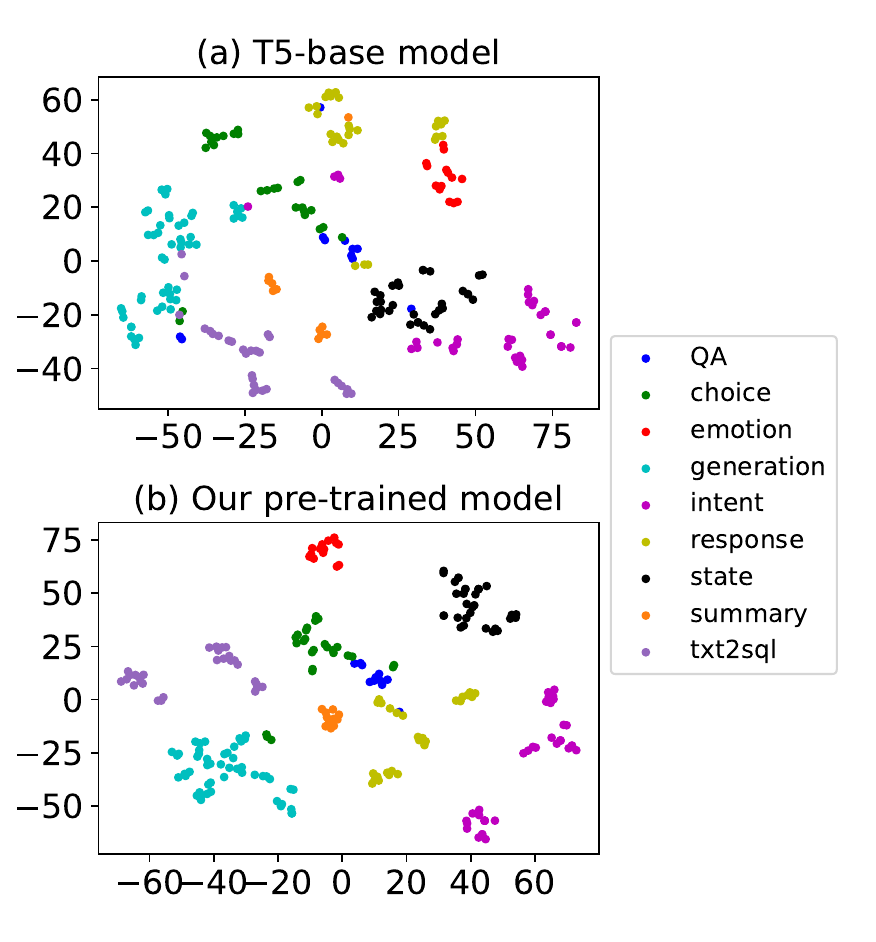}
	\caption{Prompt embeddings in the latent space using t-SNE visualization. T5-base model and our pre-trained model are used to approximate the latent space in (a) and (b) respectively.}
	\label{fig:tsne}
\end{figure}

To better understand the prompts distribution in the latent space, 
we visualize the embeddings of the prompts generated using t-SNE visualization \cite{van2008visualizing}.
As illustrated in Figure \ref{fig:tsne}, we use embeddings from language models to approximate the embeddings in the latent space, as the embeddings in the latent space are not available. The results show that our generated prompts are task-centric, yet diverse. Moreover, comparing the embeddings in our pre-trained model and T5-base model, we can see that pre-training makes the prompt embeddings of the same task cluster, meaning that the model understands the relation between tasks and prompts better after pre-training.

\subsubsection{Human Evalution}

We perform human evaluation to comprehensively evaluate the quality of the generated prompts.
We sum up three key characteristics of good prompts: task-specific, coherency and fluency, which is defined as:

\textbf{Task-specificity}: whether the prompt accurately reflects the essence of the task.

\textbf{Coherency}: whether the prompt can form coherent sequences with most of the inputs and outputs.

\textbf{Fluency}: whether the prompt itself is grammatically correct and fluent.

Experts in dialog system are asked to score 0, 1, 2 for the three metrics on the prompts generated by \methodname{} and crowdsourcing human-written prompts randomly selected from
 Prompsource \cite{bach-etal-2022-promptsource}, the average scores reported in table \ref{tab:human evaluation}. 
The results show that our generated prompts are superior to the crowdsourcing human-written prompts, improving the task-specificity, coherency, and fluency by \textbf{9.95\%}, \textbf{10.97\%}, \textbf{7.59\%} respectively. Moreover, it can be shown in the results that by modeling task in \methodname{}, the prompts generated focus on the task better, while using the input-output pairs in the automatic prompt generation procedure make the prompts generated better fit with the context, resulting in higher gain in task-specificity and coherency.

\begin{table}[t]
\caption{ Human evaluation results for prompts 
generated by \methodname{} and collected from Promptsource respectively. 
 Average scores of task-specificity, coherency, and fluency is reported.
		}
		\centering
		\resizebox{0.8\linewidth}{!}{
			\begin{tabular}{cc cccc}
				\toprule
				Method &Task-specificity  &Coherency &Fluency \\
				\midrule
		Crowdsourcing	& 1.698 & 1.687 &1.726 \\
               \methodname{} & \textbf{1.868} & \textbf{1.871} &\textbf{1.857}  \\
         \bottomrule	
		\end{tabular}}
		\label{tab:human evaluation}
	\end{table}

 \subsubsection{Results on Downstream Tasks}
 Besides human evaluation, we measure the quality of the prompts generated using the downstream finetuning results. A T5 model is finetuned on downstream tasks with the prompts generated using multi-prompt training (Eq.\ref{eq:pet-finetuning-objective}). We 
compare our automatically generated prompts with crowdsourcing prompts from Promptsource \citep{bach-etal-2022-promptsource}. Moreover, to illustrate the importance of modeling task in \methodname{}, we try to generate prompts without task-related information, i.e. the keywords for ablation study, which is the same as the method proposed in \citet{gao2021making}. The results shown in Table \ref{promt-compare} demonstrate the superiority of our automatically generated prompts over human-written prompts, improving by \textbf{2.40\%} on test accuracy. Meanwhile, modelling task in \methodname{} brings an improvement of 
 0.89\%, which shows that modeling task is beneficial for generating prompts with higher quality.

\begin{table}[t]
  \caption{Few-shot(10\%) results on BANKING77 dataset using different sets of prompts. The size of the prompt set is set to 7 as there is only 7 prompts available from Promptsource \citep{bach-etal-2022-promptsource}
		}
		\centering
		\resizebox{0.8\linewidth}{!}{
			\begin{tabular}{cc cccc}
				\toprule
				Prompts& Promptsource  &\methodname{} without task &\methodname{} \\
				\midrule
				Test Acc &80.877  &82.388 &\textbf{83.279} \\
         \bottomrule	
		\end{tabular}}
		\label{promt-compare}
	\end{table}

\section{Conclusion and Future Work}

This paper represents progress toward building high-quality dialog systems with multi-task prompt pre-training using automatically generated prompts. Based on a unified 'input-prompt-output' format, 
 we generate high-quality prompts using the proposed automatic prompt generation method 
 \methodname{} and perform multi-task prompt pre-training using the proposed  multi-prompt training mechanism, resulting in a powerful pre-trained conversation model \modelname{}.
Extensive experiments demonstrate that \modelname{} is robust to input prompts, capable of
performing various dialog-related tasks, and has strong transfer ability, particularly in low-resource scenarios. 
We hope our pre-trained model \modelname{}, as well as the collected datasets, will help researchers to build better dialog systems.
 Furthermore, since multi-task prompt pre-training is widely used in pre-training, we hope our automatic prompt generation method \methodname{}, as well as the high-quality prompts generated, will encourage the community to further explore the limits of multi-task prompt pre-training.
 




\bibliography{custom, anthology}
\bibliographystyle{acl_natbib}

 \appendix
\label{sec:appendix}
\section{Theoretical Deduction}
In this section, we give detailed theoretical deduction the superiority of our proposed \methodname{}, which is helpful in both generating high-quality prompts and the model's transfer ability on unseen tasks.

\subsection{Problem Setting}

\label{sec:model}
Given a task with input-output an instance pair $(X, Y)$, we assume that there is some 
prompt $P$ that is helpful to infer the task. Note that $P$ may be in the form of instructions, keywords, or even just the task name itself. 
Also note that in real-life the input $X$ and the prompt $P$ may not have a strict boarder, and we separate them for the convenience of discussion, assuming that $P$ contains the information relevant to the task, while $X$ contains other information:

\begin{align} 
& p(t|X,P) \approx p(t|P)
\label{prompt-assumption1}\\
&p(Y|X,P,t) \approx p(Y|X,t)
\label{prompt-assumption2}
\end{align}

We use a language model to generate output $Y$ conditioned on the input $X$ and the prompt $P$, where the task $t$ is viewed as a latent variable.
The generation probability, 
 under our latent task assumption, along with the Bayes' rule, can be written as follows:

\begin{align} 
&p(Y|X,P)\nonumber \\
&=\int_{t} p(Y|X,P,t)p(t|X,P)dt\nonumber \\
&\approx \int_{t} p(Y|X,P,t)p(t|P)dt  \quad (Eq.(\ref{prompt-assumption1})) \nonumber \\
& \propto \int_{t} p(Y|X,P,t)p(P|t)p(t)dt \nonumber \; (Bayes' \; rule)\\
&\approx \int_{t} p(Y|X,t)p(P|t)p(t)dt \quad (Eq.(\ref{prompt-assumption2}))
\label{eq:latent-task}
\end{align}

\subsection{Benefits of Using Multi-Prompt Training}
\label{multi-prompt-benefit}
We can further discuss the benefits of multi-prompt training, especially in the model's transfer ability on unseen test prompts. 
Given test prompts $P^{test}$ and $N$ training prompts $P^{i}(i=1,2, \cdots, N)$, 
by increasing the number of training prompts $N$, the minimal distance between the embedding of the test prompts and training prompts in the latent task space: $\min\limits_{i}|emb(P^{test})-emb(P^{i})|$ will reduce and $p(Y|X,P^{test})$ (the probability of generating the correct label $Y$) will increase according to Eq. \ref{eq:latent-task}.
Moreover, if the number of training prompts $N$ is large enough, the expectation of the minimal distance will reduce to any given value above 0: 
\begin{align} 
\lim_{N \to \infty}\mathbb{E}\min\limits_{i}|emb(P^{test})-emb(P^{i})| \to 0 
\end{align} 
Under this circumstance, the probability $p(Y|X,P^{test})$ 
will be 
 optimized during training, resulting in strong performance. The details of the deduction is shown in Section \ref{sec:multi-prompt theory}.
 
\subsection{Deduction of the Consistency of Multi-prompt Training}
\label{sec:multi-prompt theory}

Given the test prompt $P^{test}$, the distance between the embedding of the test prompts and the $i$th training prompts in the latent task space can be written as: 
\begin{align} 
d^{i} = |emb(P^{test})-emb(P^{i})|(i=1,2, \cdots, N) \nonumber
\end{align}

Assume that $d^{i}(i=1,2, \cdots, N)$ are independent identically distributed (i.i.d) and the supporting set of $ p(d^{i})$ is $d>0$, which means the probability $P(d) =p(d^{i}<d)>0$ for any distance $d>0$. We can prove that given any $d>0$ and any $\epsilon>0$, there exists an $N$ that ensures $p( \min \limits_{i}(d^{i})<d)>1-\epsilon$:

\begin{align} 
\forall d,\epsilon>0 \;\exists N \;s.t.\quad p( \min \limits_{i}(d^{i})<d)>1-\epsilon
\label{eq:prob_limit}
\end{align}

Theorem. \ref{eq:prob_limit} can be easily proved by calculating the probability of $p( \min \limits_{i}(d^{i})>d)$:

\begin{align} 
 p( \min \limits_{i}(d^{i})>d)&= 
\prod \limits_{i=1}^{N} (1-p(d^{i}<d))\nonumber\\
&=(1-P(d))^{N}  \quad (i.i.d)
\label{eq:prob_deduction}
\end{align}

As $P(d)>0$, we can take $N>log_{1-P(d)}\epsilon$ according to Eq.\ref{eq:prob_deduction}. Therefore the 
Theorem \ref{eq:prob_limit} is proved.

From Theorem. \ref{eq:prob_limit}, we can deduce that if $N$ is large enough, we can find a training prompt that satisfies $|emb(P^{test})-emb(P^{i})|<d$, therefore the distribution of $p(P^{test}|t)$ is close to $p(P^{i}|t)$ in the latent task space:
\begin{align} 
&KL(p(P^{test}|t)||p(P^{i}|t))< d_1\\
&KL(p(Y|X,t)p(P^{test}|t)p(t)||\nonumber\\
&p(Y|X,t)p(P^{i}|t))p(t))< d_2\\
&|\int_{t} p(Y|X,t)p(P^{test}|t)p(t) dt -\nonumber\\
&\int_{t} p(Y|X,t)p(P^{i}|t)p(t) dt|<d_3
\end{align}
$d_{1},d_{2},d_{3}$ are constant that converge to zero with the increase of the number of training prompts $N$.
Therefore the training probability of $i$th sample converges to test probability, proving the consistency of multi-prompt training:
\begin{align} 
|p(Y|X,P^{i})-p(Y|X,P^{test})| \to 0
\end{align}




\end{document}